\documentclass[conference]{IEEEtran}
\IEEEoverridecommandlockouts
\usepackage{cite}
\usepackage{amsmath,amssymb,amsfonts}
\usepackage{graphicx}
\usepackage{textcomp}
\usepackage{booktabs} 
\usepackage{graphicx}
\usepackage{amsfonts}
\usepackage{dsfont}
\usepackage{graphicx, epsfig,amsmath,amssymb,latexsym,graphics,float}
\usepackage{setspace,subfig}
\usepackage{citesort}
\usepackage{float}
\usepackage[ruled,vlined,linesnumbered]{algorithm2e}
\usepackage{booktabs}
\usepackage{lipsum}
\usepackage{multicol}
\usepackage{url}
\usepackage{algorithmic}
\def\BibTeX{{\rm B\kern-.05em{\sc i\kern-.025em b}\kern-.08em
    T\kern-.1667em\lower.7ex\hbox{E}\kern-.125emX}}
\begin{document}

\title{Harnessing Low-Fidelity Data to Accelerate Bayesian Optimization via Posterior Regularization
}

\author{\IEEEauthorblockN{Bin Liu}
\IEEEauthorblockA{\textit{School of Computer Science} \\
\textit{Jiangsu Key Lab of Big Data Security $\&$ Intelligent
Processing} \\
\textit{Nanjing University of Posts and Telecommunications}\\
Nanjing, China \\
bins@ieee.org}
}

\maketitle

\begin{abstract}
Bayesian optimization (BO) is a powerful paradigm for derivative-free global optimization of a black-box objective function (BOF) that is expensive to evaluate. However, the overhead of BO can still be prohibitive for problems with highly expensive function evaluations. In this paper, we investigate how to reduce the required number of function evaluations for BO without compromise in solution quality. We explore the idea of posterior regularization to harness low fidelity (LF) data within the Gaussian process upper confidence bound (GP-UCB) framework. The LF data can arise from previous evaluations of an LF approximation of the BOF or a related optimization task. An extra GP model called LF-GP is trained to fit the LF data. We develop an operator termed dynamic weighted product of experts (DW-POE) fusion. The regularization is induced by this operator on the posterior of the BOF. The impact of the LF GP model on the resulting regularized posterior is adaptively adjusted via Bayesian formalism. Extensive experimental results on benchmark BOF optimization tasks demonstrate the superior performance of the proposed algorithm over state-of-the-art.
\end{abstract}


\section{Introduction}
In this paper, we consider a maximization problem
\begin{equation}\label{BOF}
\underset{x\in\chi}{\max}f(x),
\end{equation}
where $f$: $\chi\rightarrow\mathbb{R}$ is a continuous real-valued function, $\chi$ a Euclidean solution domain defined in $\mathbb{R}^d$, $d$ the dimension of $x$. Suppose that there exists an $x^{\ast}\in\chi$ such that $f(x)\leq f(x^{\ast})$, $\forall x\in\chi$. The task is to find $x^{\ast}$ based on a limited number of evaluations of $f$. An evaluation consists of sampling an $x$ in $\chi$, inputting it to $f$, and then obtaining the corresponding output $y=f(x)+\epsilon$, where $\epsilon\sim\mathcal{N}(0,\sigma^2)$, at the expense of a certain amount of computational resources. We focus on cases wherein $f$ is an expensive-to-evaluate black-box function with no access to its gradient. We also assume that $f$ is smooth and can be modeled by a Gaussian process (GP). Such derivative-free expensive function optimization problems arise in many fields such as the industrial design in complex engineered systems, model selection in statistics, the hyper-parameter configuration for complex machine learning systems. BO is well recognized as a powerful framework for addressing such type of problems.

Of particular interest here is how to find or obtain a satisfactory estimate of $x^{\ast}$ with BO using as less as possible evaluations of $f$. In particular, we explore the idea of posterior regularization to accelerate the GP-UCB method of \cite{srinivas2010gaussian} by harnessing LF data. The accelerated BO algorithm (ABO) can be used for cases wherein the objective function is extremely expensive to evaluate and there is a fixed related LF data set available for exploitation. The regularization is induced by an expert fusion operator on the posterior of the BOF at each iteration of the BO procedure. An extra GP model, termed LF-GP, is trained to fit the LF data and then gets involved in the fusion operation. The impact of LF-GP on the resulting regularized posterior is dynamically adapted via Bayesian formalism.

\begin{figure}[t]
\centering
\includegraphics[width=3.5in,height=2.8in]{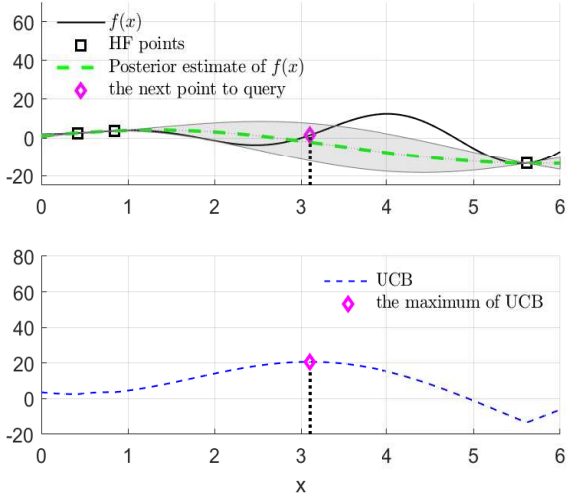}
\caption{An illustrative show of the baseline GP-UCB method on
a 1D pedagogical example: the upper panel shows the GP posterior mean and the two standard deviations band; the bottom panel shows the UCB curve obtained based on the GP posterior.}\label{fig:Demo_1D_BO}
\end{figure}

\begin{figure}[t]
\centering
\includegraphics[width=3.5in,height=2.8in]{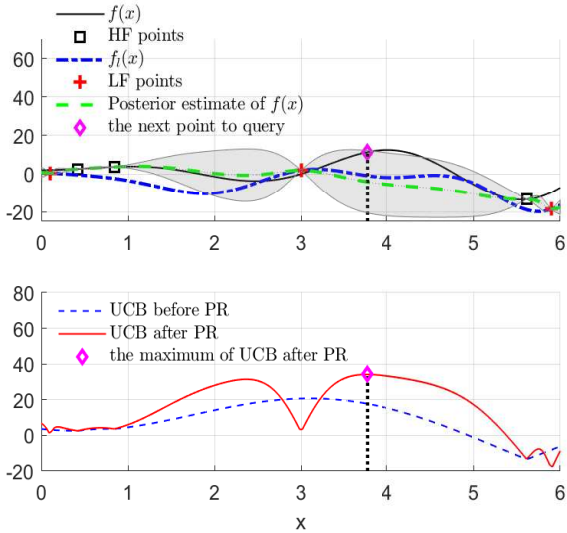}
\caption{An illustrative show of the proposed ABO algorithm on
the same example as depicted in Fig.\ref{fig:Demo_1D_BO} with 3 HF points and 3 LF points. $f_l$ is an LF approximation of $f$. The upper panel plots the regularized posterior mean and the two standard deviations band. The bottom panel shows two UCB curves obtained before and after the posterior regularization (PR), respectively.}\label{fig:Demo_1D_Proposed}
\end{figure}

The basic idea underlying the proposed ABO algorithm is illustrated in Fig.\ref{fig:Demo_1D_Proposed}. It depicts the result obtained at an iteration of ABO when applied for a 1D pedagogical case presented in subsection \ref{sec:opt_cases}. We see that ABO suggests a better next point to query than the baseline GP-UCB method whose result is plotted in Fig.\ref{fig:Demo_1D_BO}. This is due to the posterior regularization operation embedded into the ABO algorithm that helps to reveal more structural information of the BOF $f$ through exploiting LF data points. In Fig.\ref{fig:Demo_1D_Proposed}, we see that the presence of the LF point at $x=3$ makes the uncertainty band of the posterior significantly shrank at the local area of $x=3$. The UCB of the predicted $f$ therein is reduced accordingly. In contrast, the baseline GP-UCB, which is trained with only three high fidelity (HF) points, suggests evaluating $f$ at one query point near $x=3$. The resulting UCB curve of the baseline method is somewhat misleading because of the high uncertainty of the posterior estimate around $x=3$ and the structural information missing near $x=4$.
\subsubsection{Related work}
Multi-fidelity optimization has recently attracted considerable research interests. Techniques such as hierarchical partitioning \cite{sen2018multi}, hierarchical modeling \cite{qian2008bayesian} and ensemble methods \cite{peherstorfer2018survey}, are used to incorporate multiple fidelities/cheap approximations of the BOF. Most relevant to this paper is the line of work on Bayesian optimization with multi-fidelity data such as the MF-GP-UCB method in \cite{kandasamy2016gaussian} and the multi-fidelity BO (MFBO) algorithm in \cite{perdikaris2016model}. Research topics that are close to MFBO in concept include multi-information source optimization \cite{ghoreishi2018multi,poloczek2017multi}, multi-task BO \cite{swersky2013multi}, multi-output GP \cite{alvarez2011computationally,boyle2005dependent}, meta-learning based BO \cite{feurer2018scalable,feurer2014using}.

The success of the aforementioned methods requires specific assumptions to be satisfied. For instance, MFBO methods in \cite{perdikaris2016model,kennedy2000predicting,perdikaris2016multifidelity} work under a basic assumption that the relationship between $f(x)$ and $f_l(x)$ satisfies $f(x)=\rho f_l(x)+n$, where $f_l(x)$ denotes an LF approximation of $f(x)$ and $n$ a noise item. Extra operations or assumptions are usually needed to determine the value of the correlation parameter $\rho$. The hierarchical modeling approach of \cite{qian2008bayesian} requires that the data points selected for HF evaluations come from a subset of those used for LF evaluations. The MF-GP-UCB algorithm in \cite{kandasamy2016gaussian} assumes that $\|f(x)-f_l(x)\|_{\infty}$ is bounded and \emph{a priori} known.

In contrast, the application of the proposed ABO algorithm does not require any of the aforementioned assumptions to be satisfied. In ABO, a flexible expert fusion operator is embedded that automatically grasps and exploits the intrinsic correlation between the BOF and its LF counterpart. Besides, we consider a fixed LF data set $\mathcal{D}_{lf}$ for use. That said, only the HF BOF is allowed to be evaluated after starting the BO process. In contrast, in settings of most existent MFBO methods, e.g. in \cite{kandasamy2016gaussian}, new LF evaluations are allowed to perform and thus the set $\mathcal{D}_{lf}$ will be expanded accordingly. See details on the problem setup in subsection \ref{sec:problem_setup}.
\section{Preliminary}
\subsection{Problem setup}\label{sec:problem_setup}
The task is to maximize the BOF $f$ over the domain $\chi$, as formulated in Eqn.(\ref{BOF}). We search the maximizer $x^{\ast}$ or the maximum value $f^{\ast}=f(x^{\ast})$ using an algorithm that evaluates a sequence of points $x_{1:t}\triangleq\{x_1,\ldots,x_t\}, t>0$. An evaluation of $f$ at $x\in\chi$ yields an observation $y=f(x)+\epsilon$, where $\epsilon\sim\mathcal{N}(0,\sigma^2)$. There are $J$ LF query points $\mathcal{D}_l=\{x_{l,j},y_{l,j}\}_{j=1}^J$ that can be exploited, where $x_l\in\chi$, $y_l=f_l(x_l)+\varepsilon$, $\varepsilon$ denotes a zero-mean noise item. At time $t$, the algorithm chooses to query at $x_{t+1}$ based on $\{x_i,y_i\}_{i=1}^t$ and $\{x_{l,j},y_{l,j}\}_{j=1}^J$. The goal of the algorithm is to achieve as small as possible simple regret, which is defined as below
\begin{equation}\label{eqn:simple_regret}
S_t=\underset{i=1,\ldots,t}{\min}(f^{\ast}-f(x_i)).
\end{equation}
Note that we do not put any constraint on the relationship between $f$ and $f_l$ here, while our algorithm will discover and then make use of their relationship automatically and implicitly in a data-driven manner.
\subsection{Gaussian process (GP)}
A GP is a stochastic process. It is often used as a Bayesian nonparametric prior for a function $f$ defined over a space $\chi$. A GP is determined by its mean function $\mu:\chi\rightarrow\mathbb{R}$ and covariance function $\kappa:\chi^2\rightarrow\mathbb{R}$. Suppose that our prior belief on $f$ is modeled by a GP, denoted by $f\sim\mathcal{GP}(\mu,\kappa)$. This is equivalent to say that in our prior knowledge, $f(x)$ is distributed normally $\mathcal{N}(\mu(x),\kappa(x,x)), \forall x\in\chi$. Given $n$ observations $\mathcal{D}_n=\{(x_i,y_i)\}_{i=1}^n$ drawn from this GP, the posterior belief on $f$ is also a GP with an updated mean and covariance as follows
\begin{eqnarray}
\mu_n(x)&=&k^{T}(K+\sigma^2I)^{-1}Y,\\
\kappa_n(x,x')&=&\kappa(x,x')-k^{T}(K+\sigma^2I)^{-1}k',
\end{eqnarray}
where $Y=y_{1:n}$, $k,k'\in\mathbb{R}^n$ with $k_i=\kappa(x,x_i)$, $k'_i=\kappa(x’,x_i)$.
A common choice of the covariance function $\kappa$ is the squared exponential (SE) kernel, written as $\kappa(x,x')=\kappa_0\exp(-(\|x-x'\|)^2/(2h^2))$. Here $\kappa_0$ is the scale parameter that determines the extent to which $f$ could deviate from $\mu$. The bandwidth parameter $h\in\mathbb{R}_{+}$ determines the smoothness of the GP. The larger $h$ is, the smoother the samples drawn from the GP tend to be. See \cite{williams2006gaussian} for more information on the GP.
\subsection{The baseline GP-UCB method}
The GP-UCB algorithm of \cite{srinivas2010gaussian} is a typical BO method, which assigns a GP prior to $f$ and uses a UCB acquisition function to recommend new query points for evaluating $f$. At time $t$, the next point to query, $x_{t+1}$, is chosen via two steps. First, calculate a UCB of the GP as follows
\begin{equation}\label{eqn:ucb_acquisition}
\phi_t(x)=\mu_{t}(x)+\beta_t^{1/2}\sigma_{t}(x),
\end{equation}
where $\mu_{t}$ and $\sigma_{t}$ are respectively the posterior mean and standard deviation of the GP conditional on $\mathcal{D}_{t}=\{(x_i,y_i)\}_{i=1}^{t}$. Next, choose the next query point by maximizing $\phi_t$, i.e., $x_{t+1}=\underset{x\in\chi}{\max}\phi_t(x)$. This optimization can be dealt with by off-the-shelf optimization techniques, e.g., the CMA-ES method \cite{hansen2006cma}. The composites of the acquisition function, namely $\mu_{t}$ and $\sigma_{t}$ in $\phi_t$, promote exploitation and exploration, respectively. The baseline GP-UCB method is summarized in Algorithm \ref{alg:GP-UCB}. For more details on GP-UCB and other alternatives of BO methods, see \cite{shahriari2016taking}.
\begin{algorithm}[tb]
\caption{The GP-UCB Algorithm}
\label{alg:GP-UCB}
\begin{algorithmic}[1] 
\FOR{$t$=1,2,\ldots}
\STATE Train a GP to fit $\mathcal{D}_t=\{(x_i,y_i)\}_{i=1}^t$;
\STATE Find $x_{t+1}\in\chi$ by optimizing the acquisition function defined in Eqn.(\ref{eqn:ucb_acquisition}).
\STATE Sample $y_{t+1}=f(x_{t+1})+\epsilon_{t+1}$.
\STATE Augment the data $\mathcal{D}_{t+1}=\{\mathcal{D}_{t},(x_{t+1},y_{t+1})\}$.
\ENDFOR
\end{algorithmic}
\end{algorithm}
\section{The proposed ABO algorithm}
The ABO algorithm is built on the basis of GP-UCB \cite{srinivas2010gaussian}. Compared with GP-UCB, ABO is expected to be capable of employing less expensive BOF evaluations to find a satisfactory solution. The basic strategy to achieve search acceleration is to exploit an LF dataset $\mathcal{D}_{lf}=\{(x_{lf,j},y_{lf,j})\}_{j=1}^{J}$ that is assumed to be pre-available. To implement the above strategy, the key idea we adopt here is to adjust the posterior of $f$ by letting it respect predictions made by running another GP regression that uses $\mathcal{D}_{lf}$ as the training data, as shown in Fig.\ref{fig:Demo_1D_Proposed}. That says we construct two GP models in total. One is embedded in the traditional GP-UCB framework, and the other, which we term LF-GP, is trained to fit the LF data and then used for making predictions of $f$ based on the LF data. In spirit, the ABO algorithm can be regarded as an application of the posterior regularization strategy \cite{zhu2014bayesian,ganchev2010posterior} to the GP-UCB method. We develop a dynamic weighted product of experts (DW-POE) fusion operator, which generalizes the POE model of \cite{hinton2002training} by using a technique termed dynamic model averaging \cite{liu2017robust,dai2016robust,liu2011instantaneous}. The regularization is induced by the DW-POE fusion operator on the posterior. The impact of the LF-GP model on the resulting regularized posterior is adaptively adjusted via Bayesian formalism.
\begin{algorithm}[tb]
\caption{The Proposed ABO Algorithm}
\label{alg:abo}
\begin{algorithmic}[1] 
\STATE Train an LF GP to fit $\mathcal{D}_{lf}=\{(x_{lf,j},y_{lf,j})\}_{j=1}^{J}$;
\FOR{$t$=1,2,\ldots}
\STATE Train an HF GP to fit $\mathcal{D}_t=\{(x_i,y_i)\}_{i=1}^t$;
\STATE Posterior regularization: adjust the posterior of $f$, given by the HF GP, using Eqns.(\ref{eqn:reg_mu_sigma})-(11).
\STATE Find $x_{t+1}\in\chi$ by optimizing the acquisition function defined in Eqn.(\ref{eqn:reg_ucb_acquisition}).
\STATE Sample $y_{t+1}=f(x_{t+1})+\epsilon_{t+1}$.
\STATE Augment the data $\mathcal{D}_{t+1}=\{\mathcal{D}_{t},(x_{t+1},y_{t+1})\}$.
\STATE Update weights of the GP models using Eqns.(\ref{eqn:weight_predict})-(\ref{eqn:weight_update}).
\ENDFOR
\end{algorithmic}
\end{algorithm}
\subsection{The implementation of the ABO algorithm}
An implementation of the ABO is shown in Algorithm \ref{alg:abo}. First, we train an LF GP model to fit the LF data $\mathcal{D}_{lf}$. This operation is carried out off-the-shelf. Then, given any query $x\in\chi$, we invoke the LF GP model to get an LF posterior mean and standard derivation of $f(x)$, denoted by $\mu_{lf}(x)$ and $\sigma_{lf}(x)$, respectively. In the main loop of ABO, we first train an HF GP model to fit $\mathcal{D}_{t}$ at time $t$. We call it HF GP to discriminate it from the LF GP model. This HF GP model gives a posterior estimate of $f(x)$, with mean $\mu_t(x)$ and variance $\sigma_t(x)$. We adjust this posterior via the DW-POE operator, which will be described in detail in subsection \ref{sec:dw-poe}. Then, based on the adjusted posterior, we construct a UCB acquisition function as shown in Eqn.(\ref{eqn:reg_ucb_acquisition}) and find $x_{t+1}$ by optimizing the acquisition function using the CMA-ES algorithm. As shown in Eqn.(\ref{eqn:dw_poe}), a time-evolving weight $0\leq w_{lf,t}<1$ is assigned to the LF GP model when carrying out the DW-POE operator. The weight $w_{lf}$ will be adjusted along time by Eqns.(\ref{eqn:weight_predict})-(\ref{eqn:weight_update}). We give an analysis of the above algorithm design in subsection \ref{sec:analysis}.
\subsection{Dynamically weighted POE (DW-POE)}\label{sec:dw-poe}
We start by briefly describing the POE model of \cite{hinton2002training}, which is the basis of the DW-POE operator proposed for GP posterior regularization.
\subsubsection*{POE} Given multiple probability densities, $p_i(x)$, $i=1,\ldots,I$, a POE models a target probability distribution $p(x)$ as the product of $p_i(x)$'s as follows,
\begin{equation}\label{eqn:poe}
p(x)=\frac{1}{Z}\Pi_ip_i(x),
\end{equation}
where $Z$ is a normalizing constant that makes $p(x)$ a probability distribution that integrates to 1. When $p_i(x)\sim\mathcal{N}(\mu_i(x),\sigma_i^2(x)), i=1,\ldots,I$, $p(x)$ is still Gaussian, with mean and variance:
\begin{eqnarray}\label{eqn:poe_mu_sigma}
\mu(x)&=&\left(\sum_i\mu_i(x)(\sigma_i^2(x))^{-1}\right)\left(\sum_i(\sigma_i^2(x))^{-1}\right)^{-1}\\
\sigma^2(x)&=&\left(\sum_i(\sigma_i^2(x))^{-1}\right)^{-1}.
\end{eqnarray}
\subsubsection*{DW-POE for GP posterior regularization}
We generalize the POE model for GP posterior regularization. This generalized POE is termed DW-POE. The regularization is induced by the DW-POE on the posterior of $f$ given by the HF GP model. Define $p_{1,t}(x)\sim\mathcal{N}(\mu_t(x),\sigma_t^2(x))$ and $p_2(x)\sim\mathcal{N}(\mu_{lf}(x),\sigma_{lf}^2(x))$. That says we use $p_{1,t}(x)$ and $p_2(x)$ here to denote the posterior of $f$ given by the HF GP model and the LF GP model at time $t$, respectively. The regularized posterior is specified to be
\begin{equation}\label{eqn:dw_poe}
p_{reg,t}(x)\propto p_{1,t}(x)^{1-w_{lf,t}}p_2(x)^{w_{lf,t}},
\end{equation}
where $0\leq w_{lf,t}<1$ denotes a time-evolving weight assigned to the LF GP model. The time-evolving rule is specified by Eqns.(\ref{eqn:weight_predict})-(\ref{eqn:weight_update}), which will be introduced later.
Since $p_{1,t}(x)$ and $p_2(x)$ are both Gaussian, the mean and the variance of the regularized posterior can be calculated as below \cite{cao2014generalized}
\begin{eqnarray}\label{eqn:reg_mu_sigma}
\mu_{reg,t}(x)&=&\frac{\mu_t(x)w_1P_1+\mu_{lf}(x)w_2P_2}{w_1P_1+w_2P_2},\\
\sigma_{reg,t}^2(x)&=&(w_1P_1+w_2P_2)^{-1}.
\end{eqnarray}
where $w_1=1-w_{lf,t}$, $w_2=w_{lf,t}$, $P_1=(\sigma_t^2(x))^{-1}$, $P_2=(\sigma_{lf}^2(x))^{-1}$.
The UCB of the regularized GP is
\begin{equation}\label{eqn:reg_ucb_acquisition}
\phi_{reg,t}(x)=\mu_{reg,t}(x)+\beta_t^{1/2}\sigma_{reg,t}(x).
\end{equation}

The weight $w_{lf}$ is used to control the influence of the LF GP model on the regularized posterior.
The dynamic feature of $w_{lf}$ makes the DW-POE adaptable for use for different cases. Suppose a case in which the LF GP model produces a biased mean prediction with an erroneously low predicted variance. If the combination rule specified by the original POE is under use, then it can lead to a detrimental prediction of $f$, while a down-weighting of the LF GP model is beneficial for avoiding that detrimental prediction. On the other hand, when the HF GP more is more unreliable due to lack of enough training data in $\mathcal{D}_t$, an up-weighting of the LF GP model can be beneficial for providing a better prediction. The key is how to adapt the value of $w_{lf}$ smartly. We propose a data-driven approach to adapt it based on Bayesian formalism. The adaptation procedure consists of two steps. Given $w_{lf,t}$, the first step gives a prior prediction of $w_{lf,t+1}$ as follows
\begin{equation}\label{eqn:weight_predict}
\hat{w}_{lf,t+1}=\frac{w_{lf,t}^{\alpha}}{w_{lf,t}^{\alpha}+(1-w_{lf,t})^{\alpha}},
\end{equation}
where $\alpha$ is called the forgetting factor. If we let $\alpha=1$, then Eqn.(\ref{eqn:weight_predict}) reduces to $\hat{w}_{lf,t+1}=w_{lf,t}$, corresponding to case in which the posterior probability of the LF GP model at iteration $t$ is adopted as the predictive prior probability at iteration $t+1$. We set $\alpha=0.9$, instead of 1, in our experiments, to increase the impact of the new HF evaluation observation in generating the posterior at iteration $t+1$.
Upon the arrival of the new observation $y_{t+1}$, we update $w_{lf,t+1}$ as below
\begin{equation}\label{eqn:weight_update}
w_{lf,t+1}=\left\{\begin{array}{ll}
\frac{\hat{w}_{lf,t+1}\cdot l_{lf}}{\hat{w}_{lf,t+1}\cdot l_{lf}+(1-\hat{w}_{lf,t+1})\cdot l_{hf}},\;\mbox{if}\; y_{t+1}>\max(y_{1:t}) \\
\hat{w}_{lf,t+1},\quad\quad\quad\quad\quad\quad\quad\;\mbox{otherwise} \end{array} \right.
\end{equation}
where $l_{lf}=\mathcal{N}(y_{t+1}|\mu_{lf}(x_{t+1}),\sigma_{lf}^2(x_{t+1}))$ and $l_{hf}=\mathcal{N}(y_{t+1}|\mu_{t}(x_{t+1}),\sigma_{t}^2(x_{t+1}))$ are likelihoods of the GP models conditional on the observation $y_{t+1}$.
Note that the first line in Eqn.(\ref{eqn:weight_update}) is just the Bayes equation. We hope to only take advantage of high quality queries for updating weights of the GP models to avoid misleading given by low-quality queries, so we assign a prerequisite, $y_{t+1}>\max(y_{1:t})$, for updating $w_{lf,t+1}$.  As we known, a BO algorithm repeatedly executes two alternated sub-tasks: (a) approximate the objective function by a GP (exploration); (b) Search the optimum based on the learnt GP (exploitation). Eqns.(\ref{eqn:weight_predict})-(\ref{eqn:weight_update}) bias the computation to the latter sub-task. This is in spirit like the annealing mechanism adopted in simulated annealing methods.
\subsection{Computational complexity analysis}\label{sec:analysis}
We analyze the computational complexity of ABO from a completely algorithmic perspective. We do not consider the computational complexity of the BOF evaluation in this analysis. Two GP models get involved in ABO, while, one of them, the LF GP model, can be trained off-the-shelf. All required predictions given by the LF GP model can also be obtained off-the-shelf before running the main loop of the algorithm. Within the main loop, the ABO algorithm has two additional operations compared with the baseline GP-UCB method, namely the posterior regularization operation (Eqns.(\ref{eqn:reg_mu_sigma})-(11)) and the weights updating operation (Eqns.(\ref{eqn:weight_predict})-(\ref{eqn:weight_update})). They contribute a tiny amount of computation complexity. Through the above analysis, we see that ABO has the same level of computational complexity as the GP-UCB method per iteration. As will be shown in Section \ref{sec:experiment}, ABO requires less iterations than the GP-UCB method to find a good enough solution, which means that, in real applications, ABO will have smaller computational complexity in total than the baseline GP-UCB method.
\section{Experiments}\label{sec:experiment}
We compare ABO with the GP-UCB algorithm and two MFBO methods, termed MFBO-I and MFBO-II here. We considered four function optimization cases. Among the objective functions under consideration, three of them are benchmark functions used for multi-fidelity simulation in the literature. For each case, the objective function is treated as a BOF, and the maximum number of allowed evaluations of the BOF is restricted at 20. Several LF data are generated via evaluating an LF version of the BOF at query points randomly chosen from $\chi$. The performance metric adopted here is the simple regret, as defined in Eqn.(\ref{eqn:simple_regret}).

The GP-UCB method is included here as a baseline for algorithm performance comparison. MFBO-I is adapted from \cite{feurer2015initializing}, in which the HF GP model is initialized with the best query point suggested by the LF GP model that is trained to fit the LF data. MFBO-II is obtained by slightly adjusting the MF-GP-UCB method of \cite{kandasamy2016gaussian}. The only difference between MF-GP-UCB and MFBO-II lies in that the former needs to select a fidelity level for next query at each iteration, while the latter restricts the fidelity level of next query to be the highest one to fit the settings considered here.
We treat MFBO-II as a competitive posterior regularization based method, which uses a different way to regularize the posterior given by the HF GP model. Through empirical tests, we show that our proposed posterior regularization operator outperforms that used in MFBO-II.

We start by introducing the objective functions under use. We consider four functions in total. In the design of these functions, many practical issues, e.g., different relationships between the LF and the HF functions, have been considered. Therefore, we expect that experimental results revealed here can be generalized to real-life cases.
\subsection{Function optimization cases under consideration}\label{sec:opt_cases}
Here, with a slight abuse of notation, we use $x_i$ to denote the $i$th element of the vector $x$.
\subsubsection*{Case I}
First, we considered a 1D pedagogical case in which the BOF $f$ and its LF counterpart $f_l$ are specified as below
\begin{equation}
f(x)=2x^{1.2}\sin(2x)+2,\nonumber
\end{equation}
\begin{equation}
f_l(x)=0.7f(x)+(x^{1.3}-0.3)\cdot\sin(3x-0.5)+4\cos(2x)-5,\nonumber
\end{equation}
where $x\in[0,6]$.
\subsubsection*{Case II}
We then considered a 2D benchmark function used in \cite{currin1988bayesian}. It is defined as
\begin{eqnarray}\nonumber
f(x)&=&\left[1-\exp\left(-\frac{1}{2x_2}\right)\right]\\
&&\times\frac{2300x_1^3+1900x_1^2+2092x_1+60}{100x_1^3+500x_1^2+4x_1+20},\nonumber
\end{eqnarray}
where $x_i\in[0,1]$, for all $i=1,2$.
Following \cite{xiong2013sequential}, we considered an LF approximation of $f$ as below
\begin{equation}
f_l(x)=\frac{A+B+C+D}{4},\nonumber
\end{equation}
where $A=f(x_1+0.05,x_2+0.05)$, $B=f(x_1+0.05,\max(0,x_2+0.05))$, $C=f(x_1-0.05,x_2+0.05)$, $D=f(x_1-0.05,\max(0,x_2-0.05))$, and $x_i\in[0,1]$, for all $i=1,2$.
\subsubsection*{Case III}
Next we considered a 4D benchmark function, termed Park (1991) Function 1 \cite{xiong2013sequential}:
\begin{eqnarray}\nonumber
f(x)&=&\frac{x_1}{2}\left[\sqrt{1+(x_2+x_3^2)\frac{x_4}{x_1^2}}-1\right]\\ \nonumber
&&+(x_1+3x_4)\exp[1+\sin(x_3)],\nonumber
\end{eqnarray}
where $x_i\in[0,1)$, for all $i=1,2,3,4.$ Following \cite{xiong2013sequential}, we set its LF approximation to be:
\begin{equation}
f_l(x)=\left[1+\frac{\sin(x_1)}{10}\right]f(x)-2x_1+x_2^2+x_3^2+0.5.\nonumber
\end{equation}
where $x_i\in[0,1)$, for all $i=1,2,3,4.$
\subsubsection*{Case IV}
The final function considered here is Park (1991) Function 2 \cite{xiong2013sequential}:
\begin{equation}
f(x)=\frac{2}{3}\exp(x_1+x_2)-x_4\sin(x_3)+x_3.\nonumber
\end{equation}
where $x_i\in[0,1]$, for all $i=1,2,3,4.$ Its LF approximation is \cite{xiong2013sequential}:
\begin{equation}
f_l(x)=1.2f(x)-1.\nonumber
\end{equation}

Note that, in Cases I and III in Sec.4.1, the LF data is actually far from the HF data. We can see from Fig. 3 that the advantage of our method is more obvious for Cases I and III.
\subsection{Experimental results}\label{sec:exp_res}
\begin{figure}
\centering
\includegraphics[width=2.7in,height=1.8in]{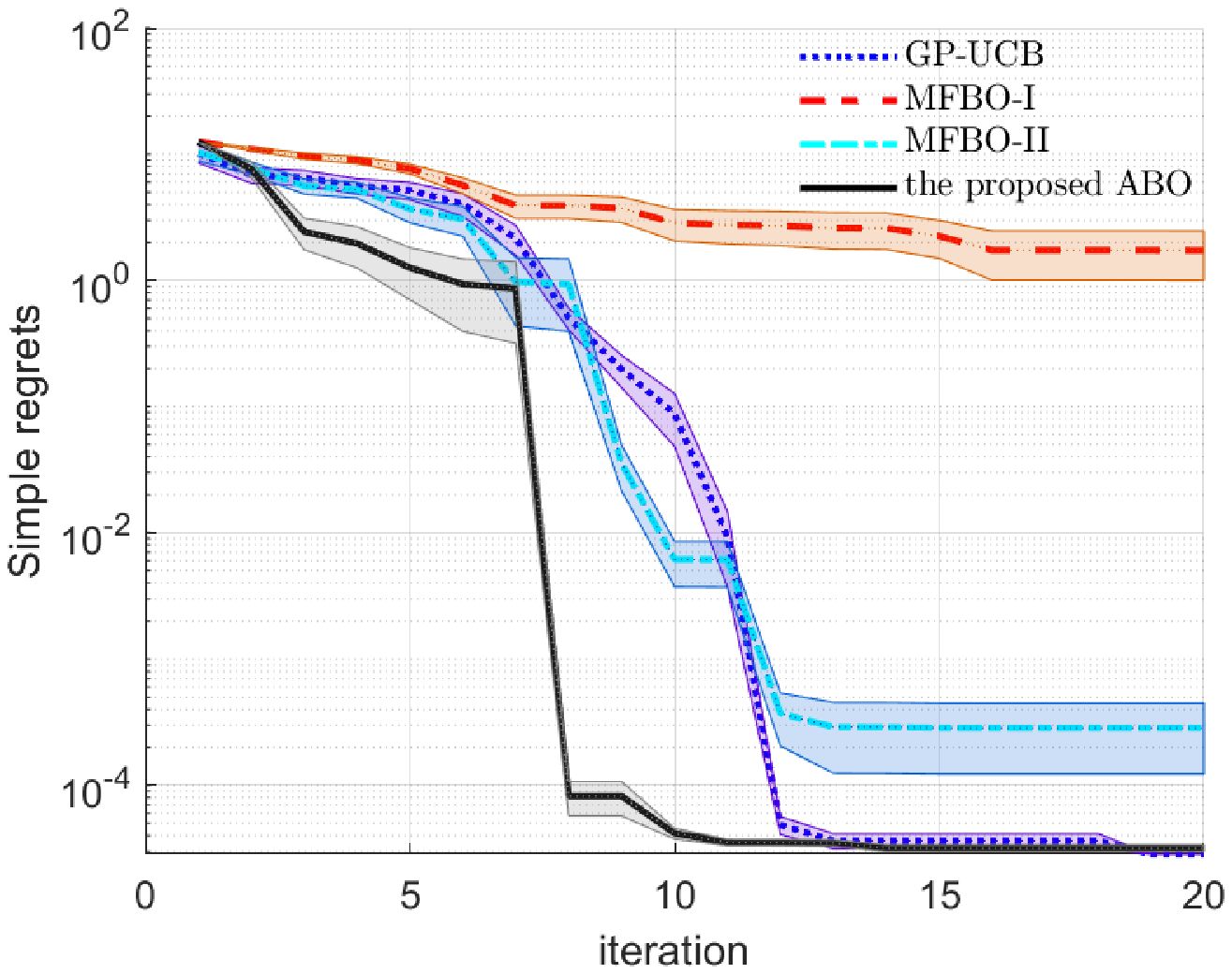}\\
\includegraphics[width=2.7in,height=1.8in]{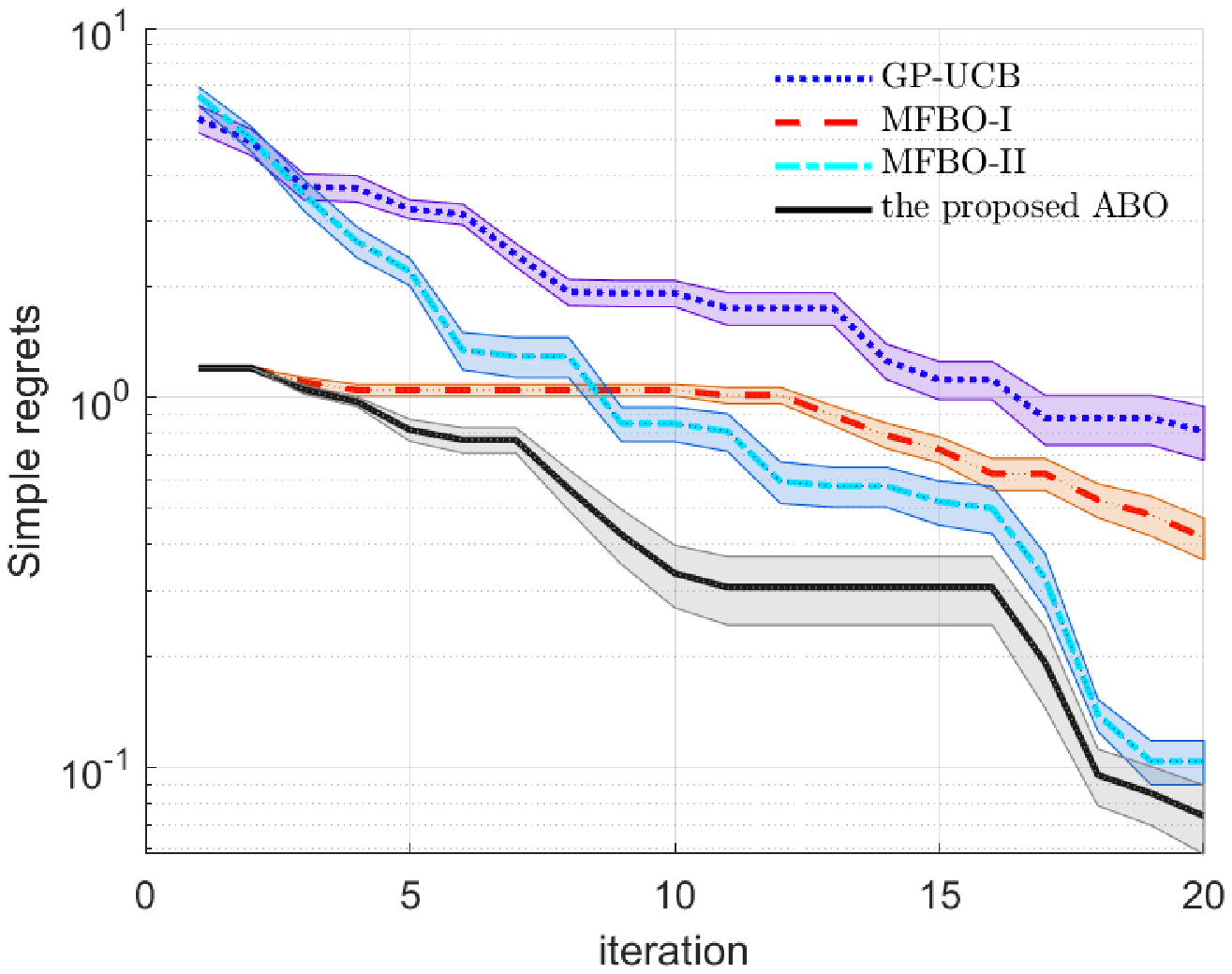}\\
\includegraphics[width=2.7in,height=1.8in]{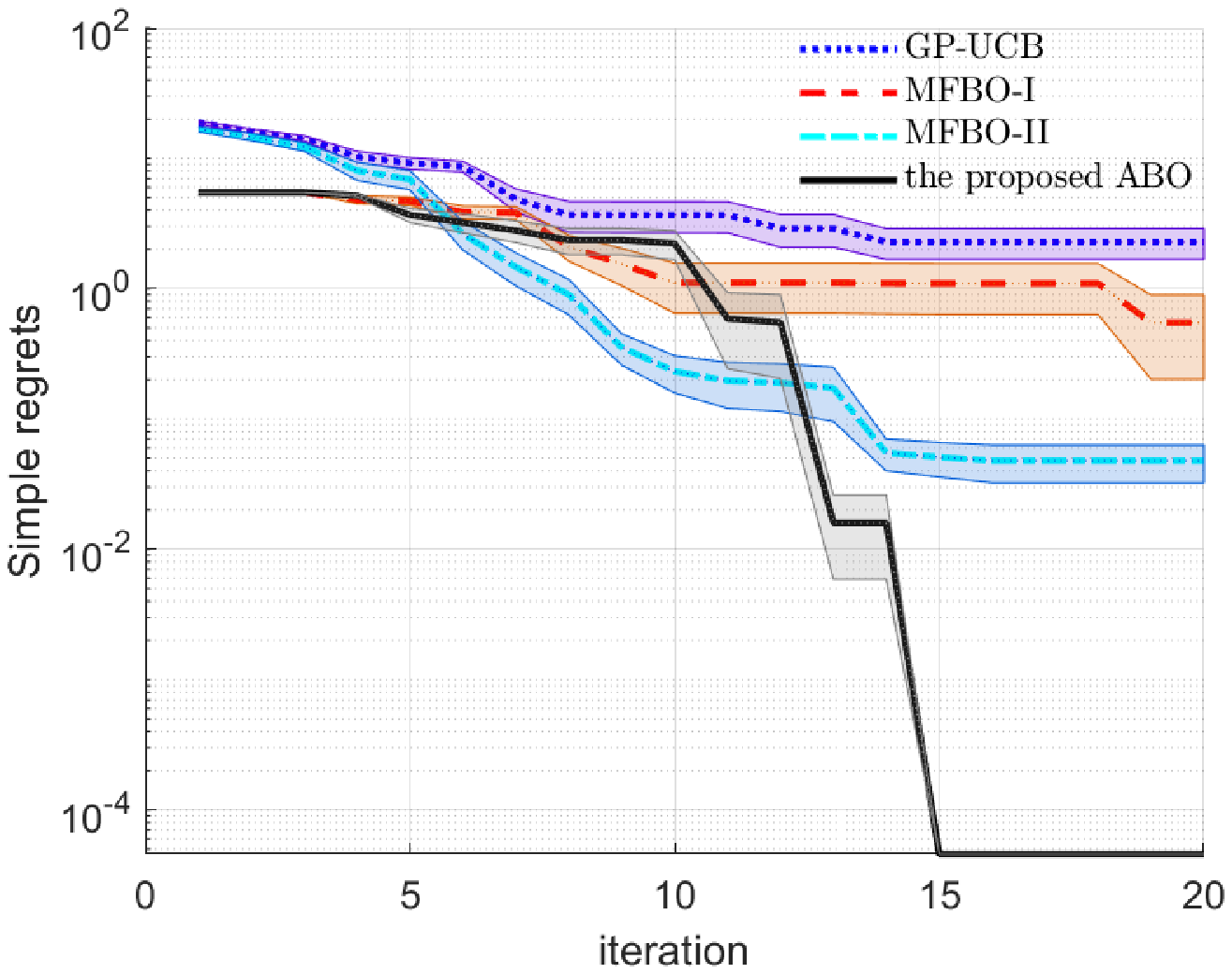}\\
\includegraphics[width=2.7in,height=1.8in]{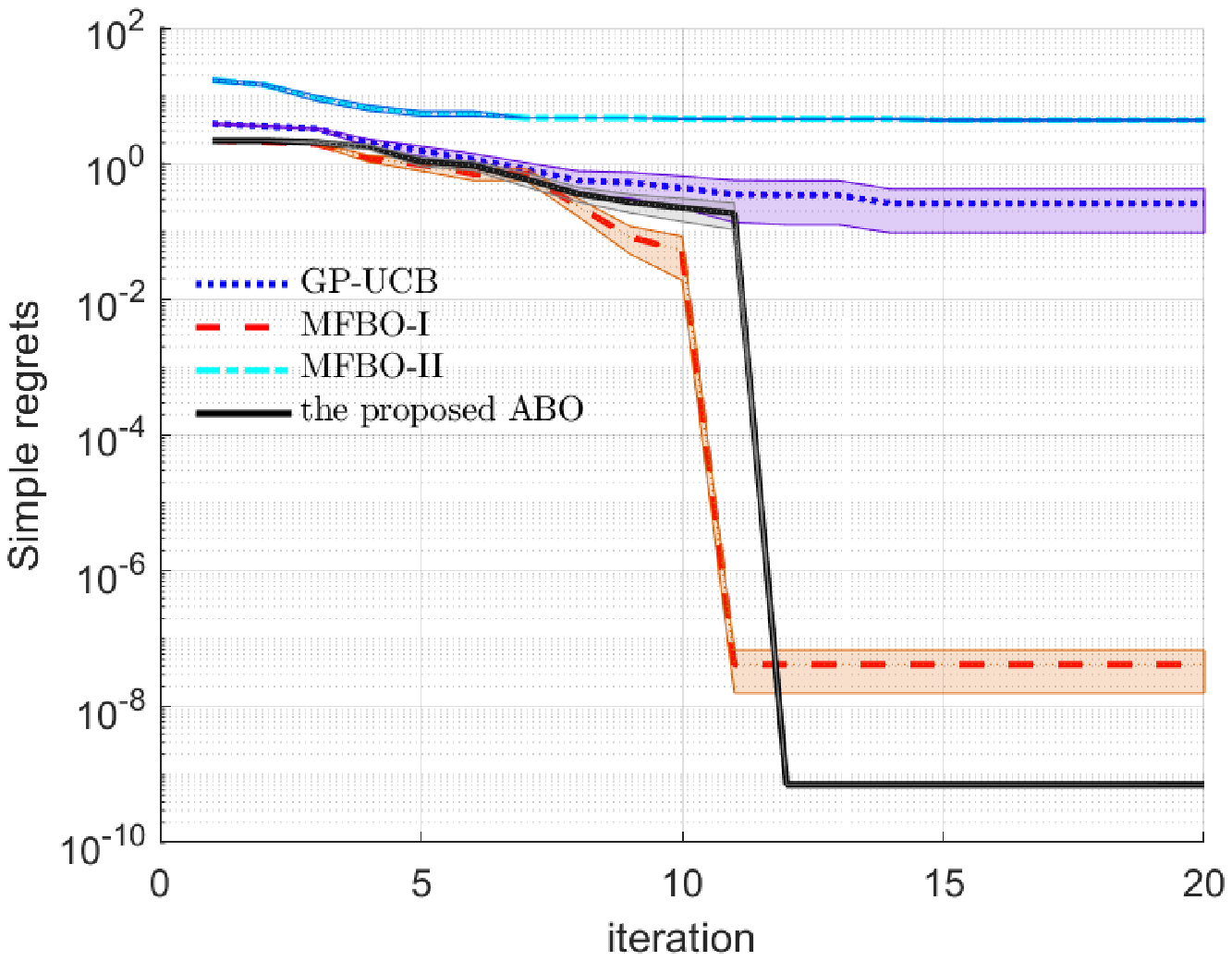}
\caption{The mean and two standard derivation band of the simple regrets over iterations. Every algorithm runs 100 times repeatedly for each case. The four panels from top to down corresponds to Case I to Case IV, respectively.}\label{fig:simple_regret}
\end{figure}

\begin{figure}
\centering
\includegraphics[width=2.7in,height=1.8in]{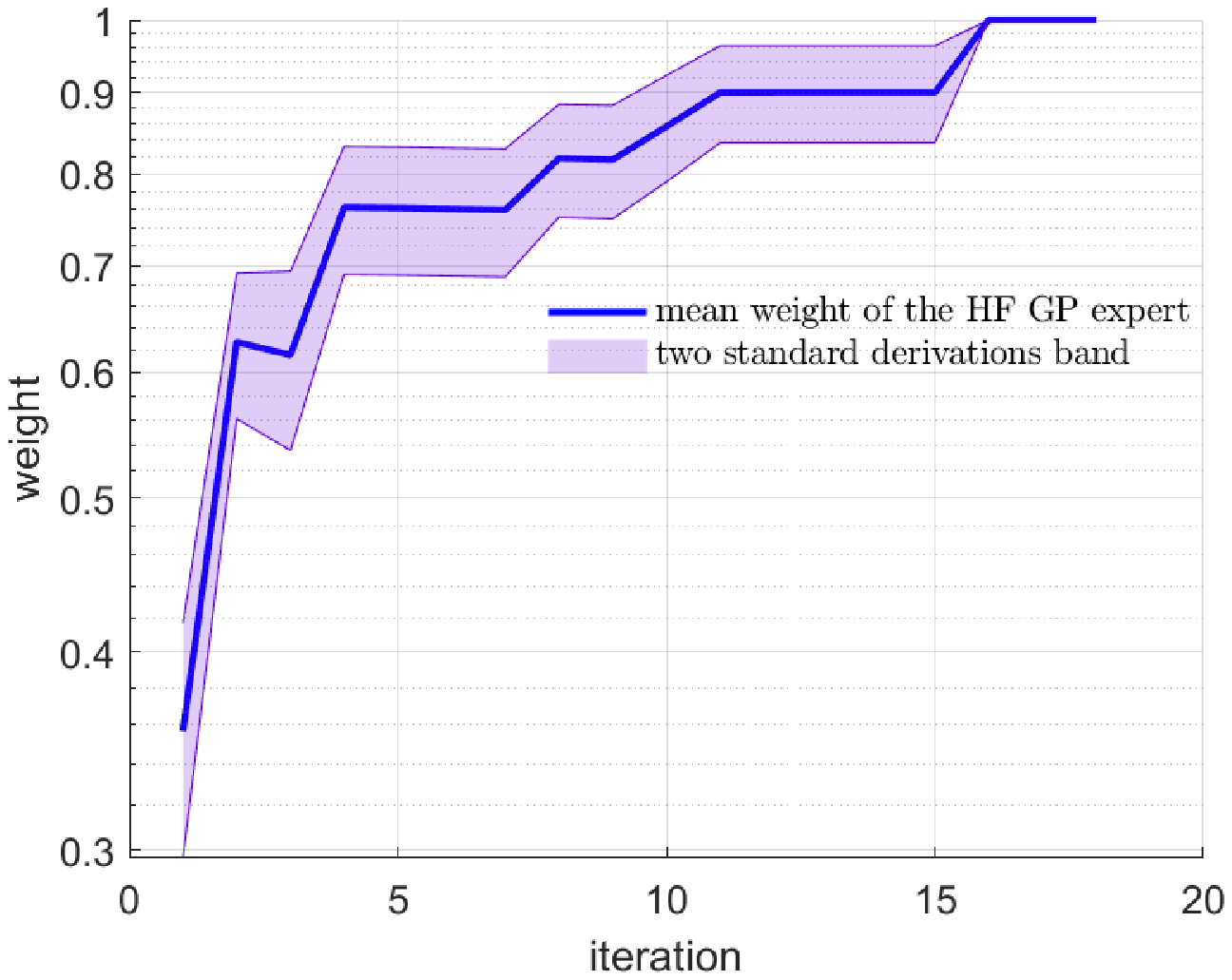}\\
\includegraphics[width=2.7in,height=1.8in]{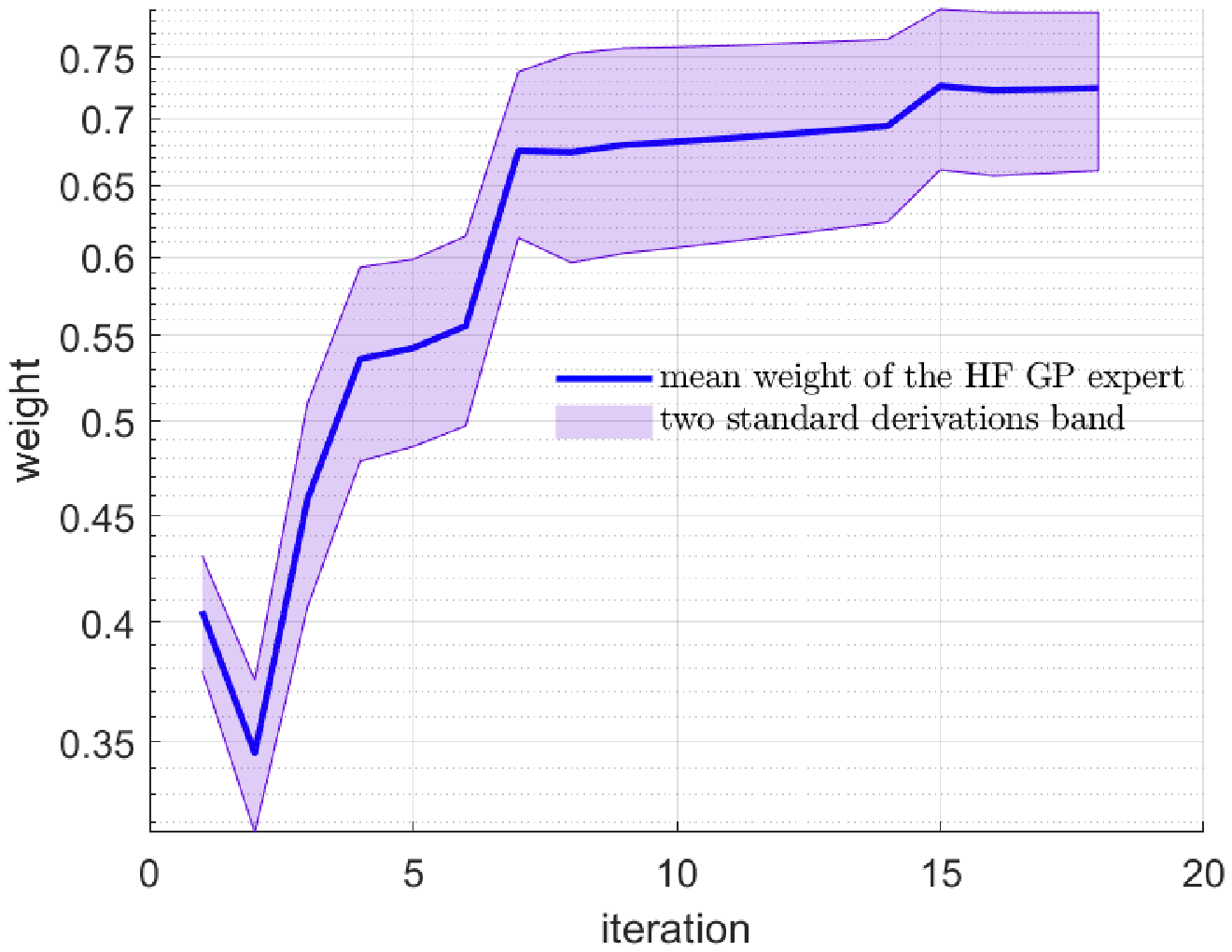}\\
\includegraphics[width=2.7in,height=1.8in]{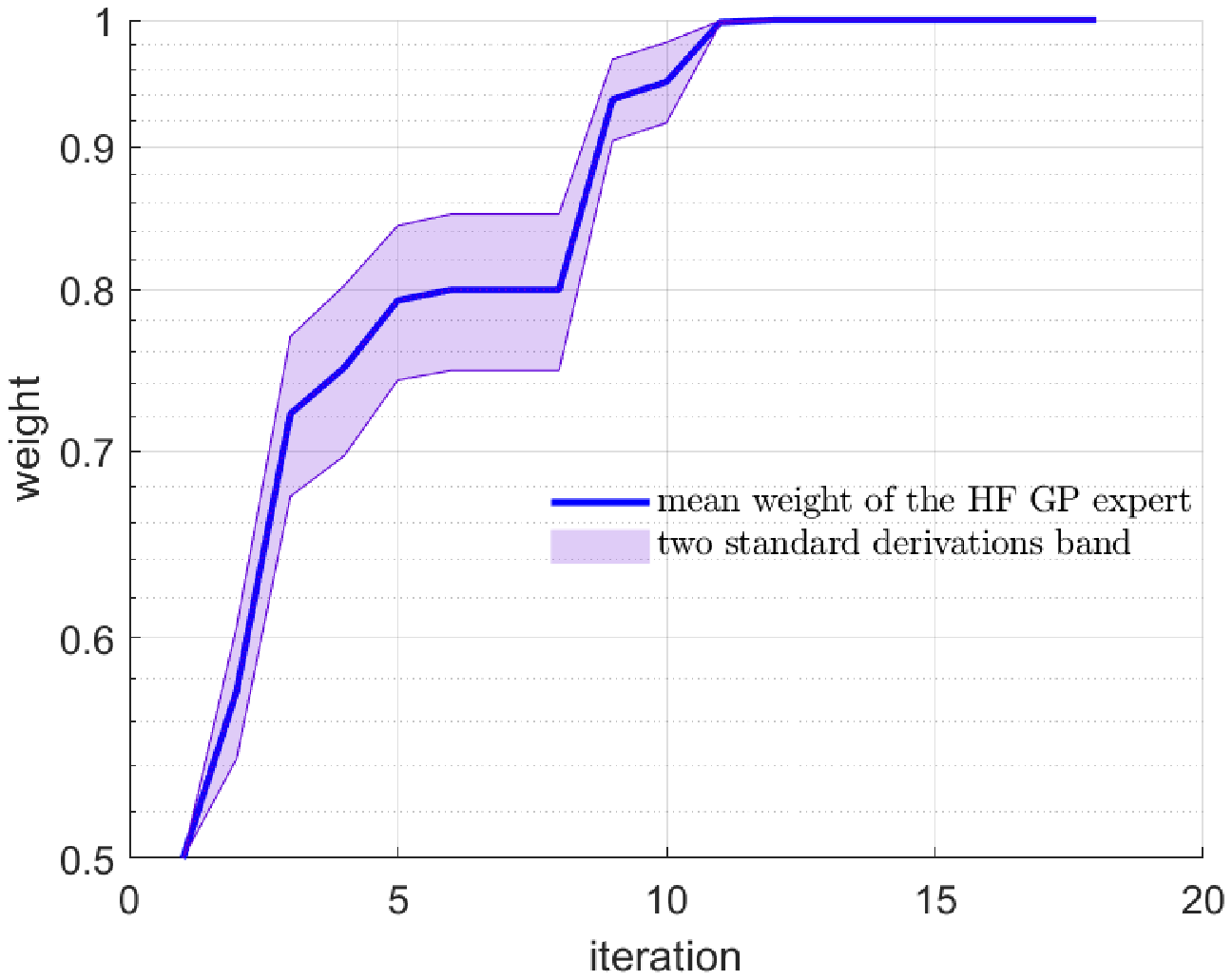}\\
\includegraphics[width=2.7in,height=1.8in]{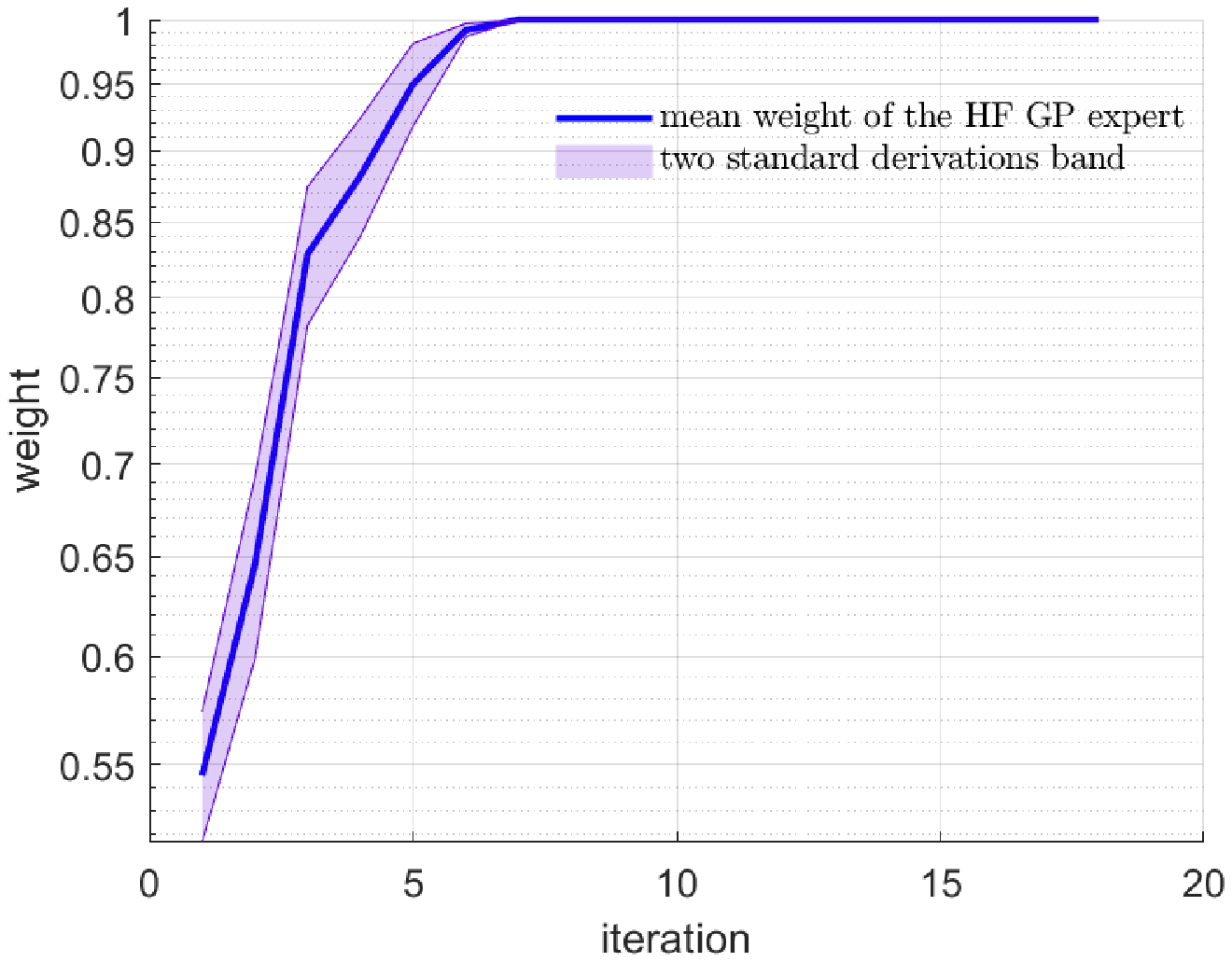}
\caption{The mean and two standard derivation band of the HF GP model's weight over 100 independent runs of the ABO algorithm. The four panels from top to down corresponds to Case I to Case IV, respectively. Note that the sum of weights of the LF GP model and the HF GP model is always 1. So the traces of the LF GP model's weight $w_{lf}$ are not plotted here.}\label{fig:weight}
\end{figure}

In the experiments, we adopted the SE kernel function and the constant mean function for GP regression, and the ``minimize.m" function in the GPML toolbox \cite{williams2006gaussian} for hyper-parameter optimization of GP models. For all algorithms considered, we adopted CMA-ES of \cite{hansen2006cma} for optimizing the acquisition function. Each algorithm is run 100 times independently to get a Monte Carlo estimate of the algorithm's performance for each case considered. The weight of the LF GP model $w_{lf}$ is initialized at 0.5 for ABO. Using Case I, we validated the mechanism of ABO for harnessing LF data to accelerate searching by visualizing an intermediate result, as shown in Figs.\ref{fig:Demo_1D_BO}-\ref{fig:Demo_1D_Proposed}. Fig.\ref{fig:simple_regret} plots simple regrets. It is shown that ABO outperforms the other methods significantly in terms of the searching speed in the first three cases. For the last case, ABO is much faster than the baseline GP-UCB method and MFBO-II, and it achieves a much smaller simple regret than MFBO-I. Fig.\ref{fig:weight} shows that the influence of the HF GP model increases along time as more HF evaluations of $f$ are performed, which conforms to our expectation.
\section{Conclusion}
In this paper, we demonstrated that LF data, which may arise from previous evaluations of an LF approximation of the BOF or
a related optimization task, can be a valuable resource for use in accelerating BO. In particular, we presented a novel algorithm design, namely ABO, for harnessing LF data to accelerate the GP-UCB algorithm of \cite{srinivas2010gaussian}.
Experimental results demonstrate that our algorithm outperforms existent state-of-the-art methods
consistently over all cases under consideration.

The basic idea underlying our method is to enable the LF data to influence the GP posterior in an automatic, data-driven and theoretically sound way.
We implemented this idea by generalizing the POE model of \cite{hinton2002training} via
Bayesian dynamic model averaging in the context of GP-UCB.
In principle, our algorithm can be regarded as an efficient approach to warm start GP-UCB by making use of related LF data.
Compared with related existent methods, the presented ABO algorithm has three major features.
First, it requires no specific assumptions on the correlation structure between the BOF and its LF approximation.
Second, the impact of the LF data is adaptively adjusted online. Specifically,
the more informative are the LF data, compared with the HF data that have already been observed, the greater is the impact of the LF data
for suggesting the next HF data point to evaluate. Lastly, the computation complexity per iteration of ABO is roughly the same as that of GP-UCB,
provided that the LF GP model has been built up beforehand.

Throughout, we make use of GP-UCB as a running example of BO methods, while the ideas presented may not be restricted to GP-UCB.
A possible future work following this line is to investigate the applicability of such ideas to accelerate other types of BO methods
and develop corresponding algorithms.
%
\bibliographystyle{IEEEbib}
\bibliography{mybib}
\end{document}